%% file: acl2023.tex
\pdfoutput=1

\documentclass[11pt]{article}

\usepackage[]{ACL2023}

\usepackage{paralist}
\usepackage{float}
\usepackage[color=blue!50,colorinlistoftodos,prependcaption]{todonotes}
\usepackage{csquotes}
\usepackage{multirow} 
\usepackage{makecell}
\usepackage{multirow}
\usepackage{longtable}
\usepackage{booktabs}
\usepackage{graphicx}
\usepackage{enumitem}
\usepackage{arydshln}
\usepackage{tcolorbox}
\usepackage{caption}

\usepackage[LGR, T1]{fontenc} 
\usepackage[main=english, greek]{babel}
\usepackage{textgreek}

\usepackage{times}
\usepackage{latexsym}

\usepackage[T1]{fontenc}

\usepackage[utf8]{inputenc}

\usepackage{microtype}
\usepackage{csquotes}

\usepackage{inconsolata}

\title{Sui Generis: Large Language Models for Authorship Attribution and Verification in Latin}

\author{Gleb Schmidt \\ Radboud University \\ Nijmegen, Netherlands \\ \texttt{gleb.schmidt@ru.nl}
        \And
        Svetlana Gorovaia \\ LEYA Lab, HSE University \\St. Petersburg, Russia\\ \texttt{sgorovaya@hse.ru}\And
        Ivan P. Yamshchikov\\ CAIRO, THWS\\W{\"u}rzburg, Germany\\
\texttt{ivan.yamshchikov@thws.de}}

\begin{document}
\maketitle
\begin{abstract}
This paper evaluates the performance of Large Language Models (LLMs) in authorship attribution and authorship verification tasks for Latin texts of the Patristic Era. The study showcases that LLMs can be robust in zero-shot authorship verification even on short texts without sophisticated feature engineering. Yet, the models can also be easily \textquote{mislead} by semantics. The experiments also demonstrate that steering the model's authorship analysis and decision-making is challenging, unlike what is reported in the studies dealing with high-resource modern languages. Although LLMs prove to be able to beat, under certain circumstances, the traditional baselines, obtaining a nuanced and truly explainable decision requires at best a lot of experimentation.
\end{abstract}

\section{Introduction}
Unlike in computational linguistics, authorship analysis in the field of digital humanities still largely relies on the complicated process of domain-specific manual feature engineering \citep{corbara_lepistola_2020, manousakis_authorship_2023, corbara_syllabic_2023, clerice_twenty-one_2023}. This is mostly due to the fact that the predictions made by machine learning models with regard to philological and historical authorship problems are expected to be contextualized within long-standing scholarly traditions with their established views on what kind of features matter in the studied material \citep{clerice_twenty-one_2023}. For this reason, although deep-learning approaches, especially those based on pre-trained language models \citep{fabien2020bertaa, rivera-soto-etal-2021-learning, ai_whodunit_2022, huertas-tato_part_2022,plutarch, wang_can_2023, huertas-tato_understanding_2024}, have recently demonstrated their reliability and high performance, even in the most sophisticated settings of virtually all authorship-related tasks\footnote{As of 2024, the following tasks have been offered at least once: Authorship Attribution, 
Authorship Clustering, 
Generative AI Detection,
Authorship Verification, 
Authorship Obfuscation
Author Profiling, and Author Diarization.} offered at PAN competitions
\footnote{https://pan.webis.de/} \citep{stamatatos_overview_2023, petropoulos_contrastive_2023, guo_contrastive_2023}. 

Pre-trained language models offer valuable insights even in challenging scenarios such as authorship analysis with limited training data, cross-discourse type verification or attribution, style change detection, and cases of stylistic mimicry where authors deliberately disguise their writing style.
Additionally, there has been a recent surge in explainable Artificial Intelligence (XAI) techniques, including feature ranking, probing, factual and counterfactual selection, attribution maps, and concept relevance propagation \cite{achtibat2023attribution}. While these methods are neither flawless \cite{setzu2024explainable} nor exhaustive, they represent a significant advancement in the field of explainability.

The linguistic \textquote{knowledge} of LLMs, acquired through training on extensive multilingual textual datasets, along with their advanced inference capabilities and their ability to provide human-like natural language explanations for their outputs inevitably raise the question of how these systems can be leveraged for philological and historical investigations.

To promote the wider adoption of large language models (LLMs) as research tools in the digital humanities, this study assesses the zero-shot performance of several publicly available, state-of-the-art LLMs --- namely GPT-4o, Gemini, Mistral, and Claude --- in authorship verification and attribution tasks. In contrast to previous studies, which have primarily worked with modern languages, our research focuses on a historical language: Latin. To the best of our knowledge, this is one of the first studies to evaluate GPT-4o’s \textquote{proficiency} in Latin and the first to report test results for the three additional models. 

Our experiments seek to evaluate the zero-shot effectiveness of LLMs in authorship verification and attribution. 
We compare their performance against traditional baselines, including classical machine learning classifiers and models based on the pre-trained Latin transformer, LaBerta. Furthermore, we investigate how variations in the quantity and nature of instructions prompted to the LLMs impact the accuracy of their predictions.

\section{Related Work} 

Authorship attribution and authorship verification are two fundamental tasks in authorship analysis. They are the most popular applications of stylometry --- the modelling of writing style using statistical methods.
Attribution, in its simplest form, aims to identify the author of a previously unseen text sample from a list of candidate authors. Verification, on the other hand, involves determining whether two given texts were written by the same person. Both tasks can vary in complexity, especially when additional challenges arise, such as cross-domain or cross-discourse type problems. 

The origins of stylometric analysis for authorship-related problems go back to the 1960s \citep{mosteller_inference_1963}. Early work in stylometry for authorship attribution relied on extraction of hand-crafted features believed to represent the writing style (word frequency, sentence length, and syntactic patterns, etc.) and Beyesian inference \citep{mosteller1984applied, holmes_authorship_1994, holmes_evolution_1998}. The study by \citet{joachims1998text} on text categorization provided a foundation for applying Support Vector Machines (SVMs) to stylometric data.
In \citeyear{burrows2002delta}, \citeauthor{burrows2002delta} suggested a distance-based technique,  which became formative for the present-day stylometry, the method now known as Burrows' Delta. Since then, various features and classification methods were used to quantify stylistic differences and estimating the likelihood of shared authorship between texts. The work of \citet{stamatatos2009survey} provides a comprehensive overview of the classical methods used in stylometric analysis. 

Since 2010s, the evolution of methodologies for solving these tasks can be traced through the overviews of PAN competitions. Until 2016, with rare exceptions, texts in authorship analysis were treated as bags-of-words \citep{stamatatos2009survey, koppel2009computational}. Research in this field revolved around various stylistic features such as word and character n-grams, sentence lengths, word and punctuation frequencies, part-of-speech (POS) tag frequencies, and POS n-grams \citep{stamatatos2013robustness}. These features were often combined with feature selection or weighting mechanisms and utilized alongside distance measures and standard classifiers like Support Vector Machines (SVM) or Naive Bayes.

The rise of the neural networks marked the shift towards closer attention to the sequential nature of the text. Convolutional and Recurrent Neural Networks (CNNs and RNNs) and later transformers have proven outperform the previous methods, particularly in cases where the writing style is more nuanced and complex \citep{shrestha2017convolutional, kestemont_overview_2018, kestemont_overview_2020}. Yet, this improvement was achieved at expense of the models's explainability.

Transformer models, such as BERT, RoBERTa, and T5, made authorship attribution and verification systems particularly robust. Since the advent of Siamese network architectures \citep{reimers_sentence-bert_2019} and the work presented by \citet{fabien2020bertaa} fine-tuning pre-trained models to solve authorship problems has de facto become a standard approach \citep{rivera-soto-etal-2021-learning, stamatatos_overview_2022, ai_whodunit_2022, huertas-tato_part_2022,plutarch, stamatatos_overview_2023, wang_can_2023, huertas-tato_understanding_2024}, although ensemble models integrating additional stylometric features \citep{fabien2020bertaa, ai_whodunit_2022} and even independent use of manually engineered features remain quite common \citep{manousakis_authorship_2023, corbara_syllabic_2023, clerice_twenty-one_2023, camps_authorship_2024}. 

Since the release of GPT-3.5 in late 2022 \citep{brown2020language} and the subsequent emergence of  GPT-4 \citep{achiam_gpt-4_2023} and a pleiade of LLMs \citep{naveed_comprehensive_2023}, there have been numerous reports of their groundbreaking performance on various research tasks relevant for the humanities \citep{karjus_machine-assisted_2023}. These tasks range from relatively simple data processing, cleaning, and structuring tasks (such as post-OCR correction, NER, and mark-up) to data augmentation and labeling \citep{tornberg_best_2024}, from semantic search to confirmatory topic analysis \citep{oiva_framework_2023}, and from text summarization and translation \citep{volk_llm-based_2024} to multimodal processing. The examples of successful applications continue to proliferate, paving the way for what \citeauthor{karjus_machine-assisted_2023} has described as \textquote{machine-assisted mixed methods} \citeyearpar{karjus_machine-assisted_2023}, which facilitate interaction with data and promise unprecedented scaling of research efforts.

Ironically, although the availability of LLMs made the detection of machine-generated text one of the most relevant real-world tasks for linguistic forensics and consequently the prevalent topic at PAN competitions \citep{bevendorff:2024}, the number of studies which explore the LLMs's own abilities to solve authorship-related problems or serve for feature extraction is rather limited so far. 

\citet{hicke_t5_2023} leveraged a pre-trained T5 model further fine-tuning for authorship identification in Early Modern English drama. \citet{patel_learning_2023} tried to bridge the gap between stylometry and language models annotating examples of writing style and creating interpretable machine-generated writing style embeddings. A somewhat comparable approach was also proposed in \citet{ramnath_cave_2024}. The model is trained using a distillation process from GPT-4-Turbo to Llama-3-8B model. First, GPT is used to produce and standardize a corpus of structured writing style descriptions. Llama is then fine-tuned to produce similar descriptions. This approach addresses the challenges of interpretability in authorship analysis by trying to establish a clear and consistent framework for it. 

An immediate source of inspiration for this study, the work by \citet{huang_can_2024}, focuses on a direct prompting of different models with authorship-related questions. The authors arrived at the conclusion that guiding the model by explicitly providing specific linguistic features to pay attention to can significantly improve the precision of the model's prediction and the quality of the analysis.

We find a compelling reason to explore the use of LLMs to be the challenge posed by sample size. Traditional machine learning methods, such as those described by \citet{eder_does_2015} and \citet{eder_short_2017}, often require samples of approximately 1000 words to achieve reliable results. Deep learning approaches typically require substantial amounts of training data, which can be difficult to obtain. In contrast, LLMs can perform effectively without extensive additional training, making them advantageous when dealing with limited or costly data resources.

While the impressive results reported in some studies \cite{fabien2020bertaa, kestemont2019overview} 
are noteworthy, it is important to stress that they were conducted using English-language datasets. Given the widespread use of English in the training data of state-of-the-art LLMs, there is a possibility that some of the datasets may overlap with the training data, potentially influencing the outcomes \cite{brown2020language}. 

The case of Latin is very different. First of all, the overall amount of available data is incomparably less. Second, it remains unknown how much of it is actually in the training data of the major LLMs. In \citeyear{burns_how_2023}, \citeauthor{burns_how_2023} evaluated the amount of Latin in the training dataset of GPT-3.5 as 339 million tokens, assuming that this number could be higher for GPT-4o and later models. Although Latin is arguably the highest-resourced of all the historical languages, the extent of the easily-available Latin dataset hardly exceeds 700 million words \citep{bamman2020latinbertcontextuallanguage} (including Neo-Latin, Latin Wikipedia, and Internet Archive), while state-of-the-art language models for this language are trained on even smaller data, a clean and high-quality subset of the extant corpus, mostly \citep{corpuscorporum, riemenschneider_exploring_2023,stroebel-roberta-base-latin-cased3,  bamman2020latinbertcontextuallanguage}. 
Nevertheless GPT-4 excels in various tasks involving Latin, such as morpho-syntactic annotation (tagging), translation from and into Latin, as well as in text summarization and paraphrasing \citep{volk_llm-based_2024}. 

To the best of our knowledge, no comprehensive study has yet been conducted on the performance of major LLMs, such as GPT-4, in the specific tasks of authorship verification and attribution in Latin. Similarly, the capabilities of other mature LLMs released after GPT-4, such as Gemini, Claude, and Mistral, have also not been thoroughly examined in this context.

For this reason, in our investigation, we tried not only to measure LLMs's performance and compare it to conventional baselines but also to study the discrepancy between different LLMs.

\section{Methodology}
We conducted a series of experiments on two tasks: authorship attribution and authorship verification. The experiments utilized direct prompting of the flagship versions\footnote{As of July-August 2024.} of four major LLMs: GPT-4o, Claude, Mistral-Large, and Gemini-1.5. However, only the authorship verification experiments involved all four models, as only GPT-4o demonstrated competitive results in the preliminary authorship attribution tests. 

All prompting was implemented in a model-agnostic manner using LangChain library. The choice was mostly dictated by the fact that this library offers a unified API to interact with many different models and facilitates crucial operations such as rate limiting, error handling (request retries), fallbacks, and, most importantly, obtaining structured output from the models.

Each run assessed the performance of a specific model on a given task within a particular setting, defined by the prompt used.

We tested three settings differed by the level of guidance the models received in addition to the default task definition: 
\begin{enumerate}
    \item BASE: the models get only a general description of the task;
    \item LIMITED: the models get a general description of the task and explicit instruction to pay attention to writing style;
    \item HIP: historically informed prompting, when the models get a general description of the task and a concise list of features to pay attention to formulated by a domain expert and anchored into the scholarly tradition. 
\end{enumerate}
Each of the aforementioned settings was tested in two different variants: basic and topic-ignorant, in which the models were explicitly instructed to avoid taking the content and theme into account. 
For the exact formulation of the prompts, see Appendix \ref{app:prompts}.

To gain further insight into the models's decision-making processes and compare their performance, we undertook two additional steps: \begin{inparaenum}[(1)] \item we investigated the influence of semantic similarity on the predictions, and \item we measured the agreement between the models
\end{inparaenum}. For the former, the texts used in our Authorship Verification experiments were vectorized using OpenAI’s text-embedding-3-large model, and a pairwise cosine similarity was calculated between them. We then computed the correlation between these similarity scores and the the models' predictions across various prompt settings. For the latter, we calculated pairwise joint probability of agreement between models, the pairwise agreement scores are presented in \ref{app:agreement}.     

\subsection{Metrics and Baselines}
To evaluate the performance of the models, we relied on accuracy, precision, recall, and F1 score. 

Furthermore, the performance of the LLMs on each of the two tasks was compared against two different baselines, (four baselines in total). For each task, one baseline features a classical machine learning approach, while the other builds upon a state-of-the-art pre-trained transformer model for Latin, LaBerta \citep{riemenschneider_exploring_2023}. For details, see the Tables \ref{tab:av_results}, \ref{tab:aa_results_250}, and \ref{tab:aa_results}. 

\section{Data} \label{sec:data}
\input{dataset}
In this study, we focus on a subset of the Patristic Sermon Textual Archive (PaSTA), a corpus of Latin homiletic literature of the Patristic era. We prefer this corpus to a seemingly more conventional Classical Roman prose for a reason. Indeed, the very nature of the genre of sermon (or homily) --- oral and written --- provides a wide spectrum of styles depending on the occasion on which sermons were delivered, the intended audience, underlying material, etc. At the very same time, the act of preaching was always framed by the scriptural and liturgical context. As the goal of the preacher was to explain the message of the Scripture, demonstrate its relevance to the everyday lives of the flocks, and make clear the symbolic and moral meaning of the sacraments and feasts of the Church, the creativity of the preacher was constantly confronted with the canons of the established genre, which suggested themes as well as discursive and rhetorical devices \citep{boodts_sermonhomiletics_2022}. Such relative thematic homogeneity of the homiletic corpus makes it a particularly interesting and complex benchmark. 

\subsection{Preparation and General Preprocessing}
For the sake of quality, the data was extracted not only from various open (\textit{Patrologia Latina} as available in the \textit{Corpus Corporum}) resources but also proprietary ones (\textit{Corpus Christianorum Series Latina}), which is why we cannot publish the full texts along with all the associated rich metadata. However, we provide all the data used in the described experiments --- the randomly sampled textual fragments with the corresponding author labels. All the data is published on GitHub \footnote{\url{https://github.com/glsch/sui_generis}.}. 

Out of the $62$ distinct authors currently represented in PaSTA, we selected $22$ authors featured in \citet{weidmann_maximus_2018}, a standard reference work to survey Latin preaching from the 3\textsuperscript{rd} to the 7\textsuperscript{th} centuries, see Table \ref{tab:dataset}. This selection covers all regions of the Late Antique Latin West and encompasses all homiletic subgenres.

Since most of the texts used in the study constitute composite entities (e.g., collections of sermons, epistles, gatherings of treatises, etc.), we first divided all the material into units (henceforth, work-units)
representing self-contained acts of preaching (e.g., \textit{sermo}, \textit{homilia}, \textit{tractatus}, epistle, \textit{dictio}). Subsequently, for different experiments, the texts were split into chunks of \emph{approximately} \begin{inparaenum}[(1)] 
    \item 250 and 
    \item 500 words.
\end{inparaenum} We opted for an oscillating chunk length to respect sentence boundaries. Therefore, some chunks are slightly longer or shorter than the target length. 

Sampling texts from the pools of chunked examples was done for each task independently. 

\subsection{Authorship Verification}

Before conducting the first authorship verification experiment, we sampled $5$ positive and $5$ negative pairs for each of the $22$ authors. This process was repeated three times, allowing us to perform each experiment with three distinct sets of pairs. While each pair was unique, the same passage could appear in multiple pairs. This yielded a balanced corpus of $660$ pairs, with $30$ pairs per author --- $15$ positive and $15$ negative. This same set of $660$ pairs\footnote{\url{https://github.com/glsch/sui_generis/blob/main/data/authorship_verification_dataset.pkl}.} was used across all subsequent authorship verification experiments, with $220$ pairs evaluated in each iteration, though the content of each iteration could vary depending on the model employed.

\subsection{Authorship Attribution}

Authorship attribution experiments were conducted using varying numbers of candidate authors: $5$, $10$, $15$, and $22$. For each of these configurations, we randomly selected the required number of authors. To ensure diversity and enhance the reliability of the results, this selection process was repeated five times, generating distinct sets of candidate authors for each iteration.

The sampling of text examples proceeded as follows. For each randomly selected author, we randomly picked two text fragments. The first fragment was designated as the query text, while the second fragment, drawn from a different work by the same author, served as the target text (i.e., the text forming a positive pair with the query). This was further supplemented with texts by other authors, which created negative pairs with the query text. The task for the model was then to match each query text with the correct target text from the provided set.

\input{tab_preprocessing}

\section{Results}
\input{av_results}
\subsection{Author Verification}

Table \ref{tab:av_results} presents the performance of each model in the tested settings, averaged over three iterations. Only two models --- GPT-4o and Claude-3.5 --- demonstrated accuracy comparable to the results reported by \citet{huang_can_2024} for English texts. Both models outperformed the LaBerta-based baseline\footnote{The model was used without fine-tuning.} in terms of accuracy, with notably high positive predictive values. Although Claude-3.5 did not outperform the baselines in terms of recall and F1 scores, its numbers were higher than those of Mistral and Gemini.

Contrarily to what was expected based on the results yielded by the so-called linguistically-informed prompt reported by \citet{huang_can_2024}, explicit philological and historical features generally deteriorated results compared to the BASE setting for all models except Gemini.

\subsection{Author Attribution}

\input{aa_results}

Tables \ref{tab:aa_results} and \ref{tab:aa_results_250} present the results of the authorship attribution task conducted on subsets of $5$, $10$, and $15$ authors, as well as on the full dataset of $22$ authors, using text fragments of $500$ and $250$ words, respectively. Only the GPT-4o model was tested for this task, as it had demonstrated the best performance in the simpler authorship verification setting.

Since multi-class classification is generally more challenging than binary classification, it is unsurprising that GPT-4o did not surpass the LaBerta baseline when the text length was sufficient ($500$ words). However, in one setting --- fragments of 250 words with $5$ authors (see Table \ref{tab:aa_results_250} in Appendix \ref{app:250}) --- GPT-4o outperformed both baselines. In all other cases, as the number of candidate authors increased, GPT-4o’s performance declined, and at a faster rate than that of the LaBerta baseline. 

Consistent with the observations from the authorship verification experiments, explicit instructions regarding philological and historical features had a negative impact on performance. Prompts with fewer constraints, such as BASE\_TOPIC\_IGNORANT or LIMITED, yielded better results. As expected, the length of text fragments had a predictable effect on prediction quality, with accuracy generally decreasing as the texts became shorter (except in the $5$-author setting). This suggests that longer fragments provide more information beneficial for authorship attribution.  

\section{Discussion}

The experiment have provided interesting insights into the capabilities of the LLMs and the way how they approach the tasks of authorship verification and attribution. 

In Authorship Verification, the strong performance of GPT-4o in the basic setting was largely anticipated due to its advanced capabilities. However, the comparable results achieved by Claude-3.5 are noteworthy, indicating its potential effectiveness in authorship verification tasks.

We were initially concerned about the high performance of the GPT-4o model in the Authorship Verification task, assuming the possibility that parts of our dataset could be simply memorized during the training and merely recalled in our experiment. The  decrease in the GPT-4o's performance observed in the Authorship Attribution task, especially with an increasing number of candidate authors, suggests that the model's decisions were guided by underlying processes other then reproducing memorized content.

In this respect, the observation that more detailed instructions, crafted by a domain expert based on scholarly tradition, actually deteriorated performance contrasts the performance of the linguistically-informed prompt used by \citet{huang_can_2024} and is perhaps particularly noteworthy. While the models are capable of detecting and describing philological features within the texts, this ability does not necessarily translate into accurate predictions. The connection between the features mentioned in HIP to the prediction is much subtler and less straightforward than that of, for example, orthography or punctuation mistakes so successfully used by \citet{huang_can_2024}. This possibly suggests that when a model can leverage its intrinsic knowledge, it achieves better results than when formal instructions are provided for tasks that are resistant to formalization \citep{ouyang2022traininglanguagemodelsfollow, liu2021pretrainpromptpredictsystematic}.

A closer examination of the models' output\footnote{All responses are available on GitHub: \url{https://github.com/glsch/sui_generis/blob/main/data/authorship_verification_responses.tsv}} highlights this issue. When explicitly instructed, the models generally perform well in identifying the specified features. For instance, they demonstrated notable \textquote{attention} to syntactical patterns such as anaphora (repetition of a word or phrase at the beginning of successive clauses), asyndeton (omission of conjunctions), polysyndeton (repetition of conjunctions), and hyperbaton (disruption of normal word order through the insertion of other words). However, in many cases, the models tend to overinterpret these features, often assuming a deterministic relationship between the presence of such patterns and the final prediction.

For example, when comparing different passages from Leander of Seville, GPT-4o generated the following description of the rhetorical devices in the two texts: \textquote{The first text uses rhetorical questions and exclamations to emphasize its points (e.g., \textquote{O infinita humilitatis documenta!}). The second text, however, relies more on a narrative and descriptive style, with extensive use of quotations from Solomon to build its argument. The rhetorical strategies differ significantly between the two texts.} Although this succinct characterization is adequate, the conclusion reached by the model is incorrect.

Similarly, GPT-4o was perplexed by a discrepancy in two different sermons by Caesarius of Arles, stating: \textquote{Text 1 employs a more complex and formal structure, with longer sentences and a higher frequency of subordinate clauses. For example, phrases like \textit{ut modestiae tuae non desit auctoritas, constantiam mansuetudo commendet, iustitiam lenitas temperet} show a sophisticated use of parallelism and balance. Text 2, while still formal, uses shorter sentences and simpler structures. It often employs direct questions and answers, such as \textit{quis est hic, et laudabimus eum?} and \textit{absit, ut desperem hic esse aliquem, immo non aliquem, sed aliquos.} This creates a more conversational tone.} Similar example can be multiplied at random.

\subsection{The Role of Semantic Similarity}

\begin{figure}[h]
    \centering
    \includegraphics[width=0.5\textwidth]{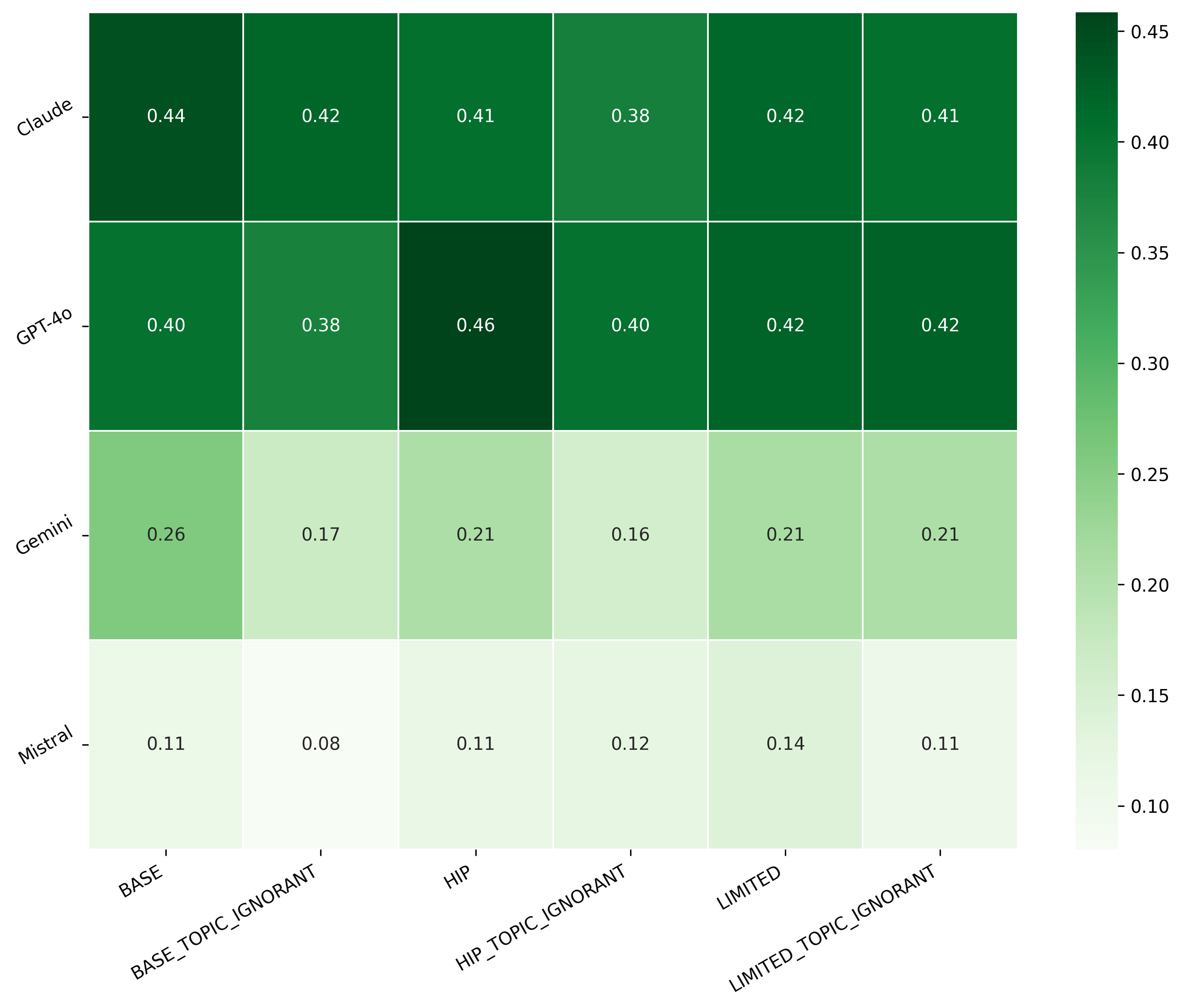}
    \caption{Cosine Similarity Correlation Heatmap by Model and Prompt}
    \label{fig:heatmap}
\end{figure}

Given the high number of such cases, we tried to analyze how semantic similarity influenced the models's decisions in the authorship verification setting. Figure \ref{fig:heatmap} represents the Pearson correlation coefficient between the cosine similarity of the prompted texts and the correctness of the model's answers. The responses of the best-performing models, GPT-4o and Claude, seem to align well with semantic similarity across various prompt settings, with only marginal variation. Even when explicitly instructed not to take the content into consideration, the models largely relied on the meaning of the texts. While writing style and semantics are inherently connected, in authorship analysis, the challenge lies precisely in discerning writing style independently of the subject matter. Our results suggest that LLMs struggle with this distinction, at least in a zero-shot setting.

LLMs are designed to follow human instructions closely, which probably explains why the settings with a lot of explicit guidance show a higher precision. However, the inherently intuitive nature of authorship analysis, especially for short texts, is not easily formalizable, which is in contradiction with what the models are trained for. 

When models are given strict prompts, they tend to follow them closely but may overinterpret features, resulting in deteriorated prediction quality. Overprompting seems to limit the models' ability to leverage their intrinsic knowledge effectively.

It is particularly clear in the case of Gemini. The model seems to have responded positively to provided instructions. With more detailed prompts, the precision of the answers increased, whereas correlation with the semantic similarity diminished. Yet, the instructions --- although formulated by a domain expert and synthesized the criteria commonly applied to authorship analysis in the field \citep{mutzenbecher_sermonum_1962,  
dolbeau_sermons_2017, weidmann_maximus_2018} ---  do not cover all possible stylistic subtleties, limiting its effectiveness. Larger models like GPT-4o and Claude benefit from less constrained prompts, allowing them to apply their extensive intrinsic knowledge more freely and leverage their capability to discern semantic similarities. We suggest that this is the reason, why the LIMITED setting, which gives provides the models with some hint to what to pay attention to and does not constrain them too much, performed that well on both tasks.

\section{Conclusion}
This study highlights the potential of large language models in performing authorship verification and attribution for Latin texts. The LLMs, particularly GPT-4o, exhibited robust performance, often surpassing traditional baselines. However, our results also highlight the challenges in steering these models' \textquote{decision-making} processes. While LLMs are capable of handling complex linguistic tasks in low-resource historical languages like Latin, there is still significant room for improvement in their interpretability and adaptability to domain-specific nuances. Enhancing their ability to disentangle style from content without relying overly on thematic similarities is crucial.

By addressing these challenges, we can unlock the full potential of LLMs in philological and historical investigations, contributing valuable tools to the fields of computational linguistics, stylometry, and the digital humanities.

\section*{Limitations}
This study and the very approach it explores have several limitations one has to keep in mind. First, in zero-shot setting we fully rely in how the models were trained by their creators, and none of the used state-of-the-art LLMs was specifically trained (or tuned) on extensive Latin datasets especially on a rather peculiar and niche task such as authorship analysis. Therefore, experimenting with it might not fully capture the potential of these models. Second, the dataset used in this study is relatively small and, as mentioned in Section \ref{sec:data}, is very peculiar from a thematic point of view. While being an interesting benchmark, it might yield observations which are difficult to generalize for texts of other epochs or genres, e.g. Latin poetry, scientific or legal prose. Third, in study, only a very superficial qualitative analysis of the output was performed. Although we present working hypotheses on the models's decision-making based on quantitative observations, the real extent of the relevance of the analysis generated by the models is yet to be determined in a close reading. We intend to investigate this in our future research. 

\section*{Ethics Statement}
This research adheres to the ethical guidelines established by the Association for Computational Linguistics (ACL). We acknowledge the limitations inherent in the use of LLMs, particularly concerning their potential biases and the ethical implications of using proprietary datasets. Care was taken to ensure that the data used did not violate any privacy or copyright concerns. The broader impact of this research is considered in terms of its contribution to the digital humanities, particularly in enhancing the tools available for studying historical texts in low-resource languages. We encourage further research that critically examines the ethical dimensions of applying LLMs to historical and cultural datasets.

\section*{Acknowledgements}
The work was supported by the ERC Starting Grant Patristic Sermons in the Middle Ages (PASSIM) and ERC Proof of concept grant ManuscriptAI. THe PI of both projects is Dr. Shari Boodts, Radboud University (Nijmegen, The Netherlands).

\bibliography{anthology, custom}
\bibliographystyle{acl_natbib}

\clearpage
\appendix
\section{Intra-model agreement}

The intra-model agreement scores reflect the reproducibility and reliability of results across models. High agreement scores, particularly with prompts incorporating topic-ignorance instruction, suggest that these prompts encourage models to make more predictions rather based on features unrelated to the subject matter of the texts.

\input{tab_agreement}

Table \ref{tab:intra_model_agreement} presents the intra-model agreement scores across different prompts for each model comparison. Generally, we observe that models demonstrate higher agreement scores when using prompts with TOPIC\_IGNORANCE instruction compared to the generic prompts. The LIMITED\_TOPIC\_IGNORANT prompt consistently yields higher agreement, especially between Claude and GPT-4o, as well as between GPT-4o and Mistral, suggesting that topic ignorance instructions positively influence intra-model consistency in predictions. Conversely, lower agreement scores are observed between Claude and Mistral, indicating that certain model-prompt pairs may interpret and respond to stylistic cues differently, even when following similar instructions.
\label{app:agreement}

\section{Experiment Settings}
Table \ref{tab:prompt_structure} summarizes the prompts used in the study.
\input{tab_settings}
\label{app:prompts}.

\section{Authorship Verification 250 words}
Table \ref{tab:aa_results_250} shows the Authorship Attribution results for fragments of $250$ words.
\input{aa_results_250}
\label{app:250}

tin patristic literature, they exhibit characteristics that point to separate authorial voices.
\end{document}

%% file: dataset.tex
\begin{table}[]
\resizebox{\columnwidth}{!}{%
\begin{tabular}{lc}
\textbf{Author}                                                    & \textbf{Word count} \\ \hline \hline
Augustine of Hippo                                          & 2,519,484  \\
Gregory the Great                                                & 794,955   \\
Origen (tr. by Rufinus)              & 385,346   \\
Caesarius of Arles                                           & 311,965   \\
Petrus Chrysologus                                              & 189,864   \\
Jerome                                                      & 178,704   \\
Optatus of Milevis                                             & 169,280   \\
Quoduultdeus                                                    & 132,160   \\
Chromatius of Aquileia                                         & 115,446   \\
Venerable Bede                                                & 114,282   \\ 
Leo the Great                                                     & 113,373   \\
Maximus of Turin (I)                                             & 73,836    \\
Gregory of Elvira                                         & 70,712    \\
Zeno of Verona                                                 & 48,077    \\
Gaudentius of Brescia                                           & 46,127    \\
Ambrose of Milan                                         & 43,118    \\
Valerian of Cimiez                                          & 31,352    \\
Basil of Caesarea (tr. by Rufinus) & 31240    \\
Priscillian of Avila                                                    & 23,165    \\
Fulgentius of Ruspa                                            & 14,804    \\
Leander of Seville                                            & 13,077    \\
Potamius of Lisbon                                           & 9,190    
\end{tabular}%
}
\caption{Dataset.}
\label{tab:dataset}
\end{table}

%% file: tab_preprocessing.tex
\begin{table*}[htbp]
\centering
\begin{tabular}{lll}
\multicolumn{1}{c}{\textbf{Experiment}}          & \multicolumn{1}{c}{\textbf{Parameter}} & \multicolumn{1}{c}{\textbf{Value}} \\ \hline
\midrule
\multirow{5}{*}{Authorship Verification} 
 & Total authors & 22 \\
 & Pairs per author & 30 (15 positive, 15 negative) \\
 & Total pairs & 660 \\
 & Repetitions & 3 \\
 & Pairs per iteration & 220 \\
 & Text length & app. 500 words \\
\midrule
\multirow{4}{*}{Authorship Attribution} 
 & Sizes of candidate author sets & 5, 10, 15, 22 \\
 & Repetitions per configuration & 5 \\
 & Texts per author & 2 (1 query, 1 target) \\
 & Pair types & 1 positive, multiple negative \\
 & Text length & app. 500 words and app. 250 words \\
\bottomrule
\end{tabular}
\caption{Sampling.}
\label{tab:preprocessing_stats}
\end{table*}

%% file: av_results.tex

\begin{table*}[h!]
\begin{minipage}{\textwidth}
\centering
\resizebox{\textwidth}{!}{%
\begin{tabular}{l|lllll}
\multicolumn{1}{c}{\textbf{Model}}          & \multicolumn{1}{c}{\textbf{Prompt/Parameters}} & \multicolumn{1}{c}{\textbf{Accuracy}} & \multicolumn{1}{c}{\textbf{Precision}} & \multicolumn{1}{c}{\textbf{Recall}} & \multicolumn{1}{c}{\textbf{F1}} \\ \hline \hline
\multirow{2}{*}{claude-3-5-sonnet-20240620} 
& BASE                     & 72 (±1\%)    & 98(±2\%)     & 45 (±4\%)  & 62 (±4\%)  \\
& BASE\_TOPIC\_IGNORANT & 68 (±1\%) &	99 (±1\%) &	37 (±4\%) &	54 (±4\%)                       \\
                                            & HIP                      & 67 (±4\%)    & 98 (±2\%)      & 34 (±10\%) & 50 (±11\%) \\
& HIP\_TOPIC\_IGNORANT &	61 (±3\%) &	100 (±0\%) &	22 (±4\%) &	36 (±5\%) \\
                                            
                                & LIMITED                      & 70 (±5\%)    & 99 (±1\%)      & 40 (±6\%) & 57 (±6\%) \\
& LIMITED\_TOPIC\_IGNORANT &	67 (±3\%) &	99 (±1\%)&	34 (±1\%) &	51 (±1\%)
\\                                \hline
\multirow{6}{*}{gemini-1.5-pro}             & BASE                     & 56 (±2\%)    & 73 (±6\%)      & 18 (±1\%)  & 29 (±2\%)   \\
                                            & BASE\_TOPIC\_IGNORANT    & 52 (±4\%)    & 57 (±11\%)    & 21 (±4\%)  & 31 (±6\%)  \\
                                            & HIP                      & 57 (±2\%)    & 77 (±4\%)      & 21 (±5\%)  & 33 (±6\%)  \\
                                            & HIP\_TOPIC\_IGNORANT     & 54 (±1\%)    & 84 (±7\%)      & 11 (±2\%)  & 19 (±4\%)  \\
                                            & LIMITED                  & 56 (±4\%)    & 84 (±9\%)     & 15 (±4\%)  & 25 (±5\%)  \\
                                            & LIMITED\_TOPIC\_IGNORANT & 55 (±7\%)    & 82 (±11\%)    & 12 (±5\%)   & 20 (±8\%)  \\ \hline
\multirow{6}{*}{gpt-4o}                     & \textbf{BASE}                     & 78 (±2\%)     & 90 (±6\%)     & 63 (±1\%)  & 74 (±2\%)  \\
                                            & BASE\_TOPIC\_IGNORANT    & 70 (±3\%)     & 95 (±1\%)     & 43 (±3\%)  & 59 (±3\%)   \\
                                            & HIP                      & 75 (±1\%)    & 88 (±0\%)      & 57 (±3\%)   & 69 (±3\%)  \\
                                            & HIP\_TOPIC\_IGNORANT     & 71 (±1\%)     & 96 (±5\%)     & 43 (±2\%)   & 59 (±3\%)  \\
                                            & LIMITED                  & \textbf{80 (±2\%)}    & \textbf{89 (±5\%)}     & \textbf{68 (±5\%)}  & \textbf{77 (±3\%)}  \\
                                            & LIMITED\_TOPIC\_IGNORANT & 70 (±3\%)    & 95 (±2\%)     & 43 (±1\%)   & 59 (±1\%)  \\ \hline
\multirow{6}{*}{mistral-large-latest}       & BASE                     & 56 (±3\%)     & 54 (±3\%)     & 76 (±5\%)   & 63 (±4\%)  \\
                                            & BASE\_TOPIC\_IGNORANT    & 54 (±3\%)     & 54 (±4\%)     & 56 (±3\%)  & 55 (±3\%)   \\
                                            & HIP                      & 54 (±6\%)    & 53 (±7\%)     & 75 (±3\%)    & 62 (±6\%)  \\
                                            & HIP\_TOPIC\_IGNORANT     & 54 (±5\%)    & 53 (±7\%)     & 63 (±5\%)  & 58 (±6\%)  \\
                                            & LIMITED                  & 56 (±1\%)    & 54 (±1\%)     & 76 (±1\%)  & 63 (±1\%)  \\
                                            & LIMITED\_TOPIC\_IGNORANT & 53 (±6\%)    & 53 (±9\%)     & 47 (±7\%)   & 50 (±7\%) \\ \hline
\multirow{1}{*}{TF-IDF + Random Forest} 
&  char, ngram\_range=2,9, max\_features=5000                    & 58    & 59      & 61   & 60  \\
\multirow{1}{*}{LaBerta + Mean pooling + Cosine similarity} 
&                      & 69    & 54      & 93   & 68  \\ \hline
\end{tabular}%
}
\caption{Results for Authorship Verification task on full dataset (22 authors, 5 positive and 5 negative pairs per author in each iteration).}
\label{tab:av_results}
\end{minipage}
\end{table*}

%% file: aa_results.tex
\begin{table*}[t]
\centering
\resizebox{\textwidth}{!}{%
\begin{tabular}{l|l|ll|ll|ll|ll}

\multicolumn{1}{c}{\textbf{Model}}          & \multicolumn{1}{c}{\textbf{Prompt/Setting}} & \multicolumn{2}{c}{\textbf{5 Authors}} & \multicolumn{2}{c}{\textbf{10 Authors}} & \multicolumn{2}{c}{\textbf{15 Authors}} & \multicolumn{2}{c}{\textbf{22 Authors}}\\ \hline \hline
\multicolumn{1}{c}{}          & \multicolumn{1}{c}{}   & \multicolumn{1}{c}{\textbf{Acc.}} & \multicolumn{1}{c}{\textbf{F1}} &  \multicolumn{1}{c}{\textbf{Acc.}} & \multicolumn{1}{c}{\textbf{F1}} &  \multicolumn{1}{c}{\textbf{Acc.}} & \multicolumn{1}{c}{\textbf{F1}} &  \multicolumn{1}{c}{\textbf{Acc.}} & \multicolumn{1}{c}{\textbf{F1}}\\ \hline 

\multirow{6}{*}{GPT-4o}                     & BASE                     & 48	& 37	 & 32	& 24 & 37	& 28	& 21   & 13
\\ 
& \textbf{BASE\_TOPIC\_IGNORANT}    & \textbf{68}	& \textbf{62}	  & \textbf{36}	& \textbf{27} & 37	& 28 & 21 & 12
\\
& HIP                      & 56	& 48	 & 32	& 23 & 35	& 28   & \textbf{21} & \textbf{13}
\\
& HIP\_TOPIC\_IGNORANT     & 44	& 36	& 32	& 22	& \textbf{39}	& \textbf{29}  & 17 & 8
\\
& LIMITED                  & 52	& 42	 & 28	& 19 & 29	& 22	& 20  & 11
\\
& LIMITED\_TOPIC\_IGNORANT & 56	& 46	& 34	& 25 & 35	& 26    & 14 & 7
\\
\hline
\multirow{1}{*}{TF-IDF}  &     & 44  & 37  & 26	& 19 & 12	& 7	& 6    & 4
\\
\multirow{1}{*}{LaBerta + Mean pooling + Cosine} &   & \textbf{72}	& \textbf{65}	&  42	& 34 & \textbf{41}	& \textbf{35} & \textbf{36}	& \textbf{29}	
\\

\end{tabular}%
}
\caption{Results for Authorship Attribution task on subsets of 5, 10, 15 and 22 (full dataset) authors with fragments of 500 words in terms of Accuracy and Weighted F1. The results of GPT-4o model are compared with several baseline pre-trained models.}
\label{tab:aa_results}
\end{table*}

%% file: tab_agreement.tex
\begin{table*}[h!]
\centering
\resizebox{\textwidth}{!}{%
\begin{tabular}{l|llllll}
\multicolumn{1}{c|}{\textbf{Prompt}} & \multicolumn{1}{c}{\textbf{Claude vs Gemini}} & \multicolumn{1}{c}{\textbf{Claude vs GPT-4o}} & \multicolumn{1}{c}{\textbf{Claude vs Mistral}} & \multicolumn{1}{c}{\textbf{Gemini vs GPT-4o}} & \multicolumn{1}{c}{\textbf{Gemini vs Mistral}} & \multicolumn{1}{c}{\textbf{GPT-4o vs Mistral}} \\ \hline \hline
BASE & 79.09 & 81.06 & 44.85 & 68.48 & 38.03 & 50.00 \\ \hline
HIP & 79.09 & 81.06 & 44.85 & 68.48 & 38.03 & 50.00 \\ \hline
LIMITED & 80.91 & 80.61 & 43.79 & 66.67 & 36.52 & 51.97 \\ \hline
BASE\_TOPIC\_IGNORANT & 72.73 & 85.61 & 53.03 & 69.85 & 46.82 & 54.09 \\ \hline
HIP\_TOPIC\_IGNORANT & 88.03 & 84.85 & 44.55 & 78.64 & 42.73 & 49.09 \\ \hline
LIMITED\_TOPIC\_IGNORANT & 81.97 & 87.12 & 56.97 & 78.48 & 53.48 & 55.61 \\ \hline
\end{tabular}%
}
\caption{Intra-model agreement scores across different prompts for model comparisons.}
\label{tab:intra_model_agreement}
\end{table*}

%% file: tab_settings.tex
\begin{table*}[h!]
\centering
\resizebox{\textwidth}{!}{%
\begin{tabular}{ll}
\multicolumn{2}{c}{\textbf{Prompt structure}}                   \\ \hline \hline
\multicolumn{2}{c}{\textbf{System message}} \\
\multicolumn{2}{c}{\parbox[t]{0.5\hsize}{You are an experienced philologist who specializes in post-Classical Latin and has a deep knowledge of Latin patristic literature. Your task is to verify the authorship of texts.}}\\
\multicolumn{2}{c}{\textbf{Taske definition}} \\
\multicolumn{1}{c|}{\textbf{Authorship Verification}} & \multicolumn{1}{|c}{\textbf{Authorship Attribution}} \\
\multicolumn{1}{l|}{
    \parbox[t]{0.5\hsize}{You will be given a pair of texts, and you will have to analyze them in order to decide whether they are written by the same author or not. Importantly, you do \textbf{not} have to guess who the author is, but only decide whether the provided texts are likely to be written by the same person or not.}
} & 
\multicolumn{1}{|l}{
    \parbox[t]{0.5\hsize}{Given a set of texts with known authors and a query text, determine the author of the query text.}
} \\ \hdashline
\multicolumn{2}{c}{\textit{Optional parameter}}                 \\
\multicolumn{2}{c}{\parbox[t]{0.5\hsize}{\textbf{TOPIC IGNORANCE} As the texts are thematically similar and all of them feature religious, theological, and philosophical content, you should disregard in your decision the topic and content (an additional instruction which can be prepended to any other).}}                    \\
\multicolumn{2}{c}{\textbf{Guidance levels}}                    \\
\multicolumn{2}{l}{\parbox[t]{\hsize}{\textbf{BASE} \textit{Task definition only, no further guidance provided except for optional TOPIC IGNORANCE.}}}                            \\
\multicolumn{2}{l}{\parbox[t]{\hsize}{\textbf{LIMITED} Base your reasoning on the analysis of the writing style of the input texts.}}                            \\
\multicolumn{2}{l}{\parbox[t]{\hsize}{\textbf{HIP} (historically-informed prompt)  Carry out your analysis by examining the philological and historical elements of the writing style found in the input texts. Consider, but do not limit your analysis to, the following features:
        \begin{itemize}[noitemsep,topsep=0pt]
            \item Morphology: affixes, declination, and verbal endings
            \item Syntax: sentence structure, use of tenses and moods
            \item Rhetorical figures: tropes and figures of speech which alter the ordinary meaning or order of words to produce rhetorical effects or rhythmical patterns.
            \item The use of the Bible: how biblical quotations are introduced, framed, and/or connected to each other
            \item Vocabulary of the text: compound and modal verbs; the words authors use to make evident the structure of the argument as well as various function words (conjunctions, pronouns, interjections, and particles) and the so-called hapax legomena (rare word and expressions)
            \item The tone of the text (moralizing, philosophical, exegetical, high-flown, affectionate, chunky, simplistic, etc.)
        \end{itemize}}} \\ \hline
\multicolumn{2}{c}{\textbf{Human message}} \\
\multicolumn{1}{c|}{\textbf{Authorship Verification}} & \multicolumn{1}{|c}{\textbf{Authorship Attribution}} \\
\multicolumn{1}{l|}{\parbox[t]{0.5\hsize}{
        \begin{itemize}[noitemsep,topsep=0pt]
            \item Text 1
            \item Text 2
        \end{itemize}}}   &     \multicolumn{1}{|l}{\parbox[t]{0.5\hsize}{
        \begin{itemize}[noitemsep,topsep=0pt]
            \item Query text
            \item Texts of candidate authors
        \end{itemize}}}                     \\

\end{tabular}%
}
\caption{Prompt structure and experiment settings.}
\label{tab:prompt_structure}
\end{table*}

%% file: aa_results_250.tex
\begin{table*}[h!]
\resizebox{\textwidth}{!}{%
\begin{tabular}{l|l|ll|ll|ll|ll}

\multicolumn{1}{c}{\textbf{Model}}          & \multicolumn{1}{c}{\textbf{Prompt/Setting}} & \multicolumn{2}{c}{\textbf{5 Authors}} & \multicolumn{2}{c}{\textbf{10 Authors}}    & \multicolumn{2}{c}{\textbf{15 Authors}}  & \multicolumn{2}{c}{\textbf{22 Authors}}\\ \hline \hline
\multicolumn{1}{c}{}          & \multicolumn{1}{c}{}   & \multicolumn{1}{c}{\textbf{Acc.}} & \multicolumn{1}{c}{\textbf{F1}} &  \multicolumn{1}{c}{\textbf{Acc.}} & \multicolumn{1}{c}{\textbf{F1}} &  \multicolumn{1}{c}{\textbf{Acc.}} & \multicolumn{1}{c}{\textbf{F1}}  &  \multicolumn{1}{c}{\textbf{Acc.}} & \multicolumn{1}{c}{\textbf{F1}} \\ \hline 

\multirow{6}{*}{GPT-4o}                     & BASE                     & 64	& 55	  & 30 & 21 & 31   & 23    & 18	& 11\\ 
& BASE\_TOPIC\_IGNORANT    & 56	& 47   & 30 & 24 & 25   & 17    & 18	& 12
\\
& HIP                      & 60	& 50    & 28 & 19 & 21  & 14    & 22	& 15
\\
& HIP\_TOPIC\_IGNORANT     & 52	& 40    & 28 & 19   & 21    & 15    & 17	& 11
\\
& LIMITED                  & \textbf{68}   & \textbf{61} & 30   & 27    & 24    & 16 & 23	& 17
\\
& LIMITED\_TOPIC\_IGNORANT & 68	& 59    & 32    & 22   & 28   & 19    & 17	& 11
\\
\hline
\multirow{1}{*}{TF-IDF}  &     & 20	& 15  & 14  & 11 & 13   & 10    & 8	& 5
\\
\multirow{1}{*}{LaBerta + Mean pooling + Cosine} &  & 48  & 39  & \textbf{66} & \textbf{57} & \textbf{40}    & \textbf{31}    &  \textbf{39}	& \textbf{33}
\\

\end{tabular}%
}
\caption{Results for Authorship Attribution task on subsets of 5, 10, 15 and 22 (full dataset) authors with fragments of 250 words  in terms of Accuracy and Weighted F1. The results of GPT-4o model are compared with several baseline pre-trained models.}
\label{tab:aa_results_250}
\end{table*}